\documentclass{l4dc2025}

\usepackage{amsmath}
\usepackage{cleveref}

\usepackage{times}
\usepackage{subcaption}
\usepackage{graphicx}
\usepackage{latexsym}
\usepackage{paralist}

\usepackage{wrapfig}

\newcommand{\labelhighlight}[1]{(\textbf{#1})}

\title[On Foundation Models for Dynamical Systems]{On Foundation Models for Dynamical Systems \\from Purely Synthetic Data}

\author{%
 \Name{Martin Ziegler} \Email{martin.ziegler@dsme.rwth-aachen.de}
 \\
 \Name{Andres Felipe Posada-Moreno} \Email{andres.posada@dsme.rwth-aachen.de}
 \\
 \Name{Friedrich Solowjow} \Email{solowjow@dsme.rwth-aachen.de}
 \\
 \Name{Sebastian Trimpe} \Email{trimpe@dsme.rwth-aachen.de}\\
 \addr Institute for Data Science in Mechanical Engineering (DSME), RWTH Aachen University, Germany
}

\begin{document}

\maketitle

\begin{abstract}
    Foundation models have demonstrated remarkable generalization, data efficiency, and robustness properties across various domains. In this paper, we explore the feasibility of foundation models for applications in the control domain.
    The success of these models is enabled by large-scale pretaining on Internet-scale datasets. These are available in fields like natural language processing and computer vision, but do not exist for dynamical systems. 
    We address this challenge by pretraining a transformer-based foundation model exclusively on synthetic data and propose to sample dynamics functions from a reproducing kernel Hilbert space.
    Our pretrained model generalizes for prediction tasks across different dynamical systems, which we validate in simulation and hardware experiments, including cart-pole and Furuta pendulum setups. Additionally, the model can be fine-tuned effectively to new systems to increase performance even further.
    Our results demonstrate the feasibility of foundation models for dynamical systems that outperform specialist models in terms of generalization, data efficiency, and robustness.
\end{abstract}


\section{Introduction}

\begin{figure}[tb]
    \centering
    \includegraphics[width=0.75\textwidth]{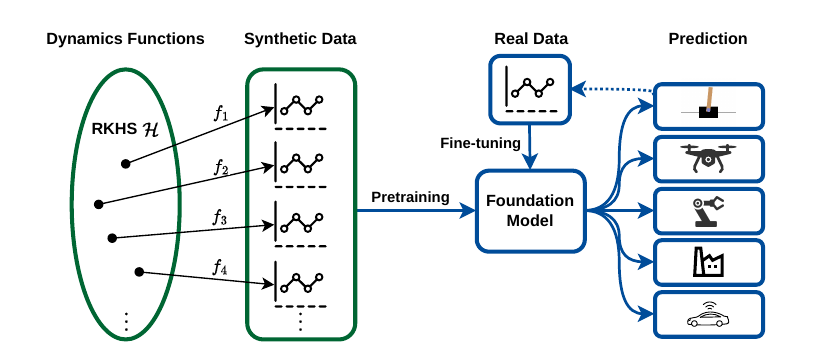}
    \caption{Proposed approach for a foundation model that predicts future states of dynamical systems. The model is pretrained purely on synthetic data (green). We propose to sample dynamics functions from an RKHS $\mathcal{H}$ and creating trajectory data by evolving the system through time. The foundation model is capable to zero-shot unseen systems in simulation and hardware (blue). Further, it can quickly be fine-tuned (dashed) to specific systems to improve the performance even further.}
    \label{fig:introduction:concept}
\end{figure}

Foundation models are trained on vast amounts of general data, allowing them to acquire extensive knowledge about a domain. 
This extensive pretraining enables them to easily adapt to new downstream tasks via no or little fine-tuning, often outperforming specialist models trained only for specific tasks \citep{bommasani2021opportunities,mann2020language}. 
Specifically, foundation models have demonstrated unprecedented capabilities with regard to the following properties:
\begin{compactenum}
    \item[\labelhighlight{P1}] \textbf{Generalization}: The ability to perform well on a wide range of tasks, including those not explicitly trained on, by applying learned knowledge to new, unseen situations effectively \citep{yuan2023powerfoundationmodels,bommasani2021opportunities,liu2024few}.
    \item[\labelhighlight{P2}] \textbf{Data efficiency}: Using a pretrained foundation model on a new task requires small or no amounts of task-specific data,
    indicating the model's capacity to learn and adapt quickly without requiring extensive additional training data \citep{mann2020language,schick2020s,liu2024few}.
    \item[\labelhighlight{P3}] \textbf{Robustness}: 
    Continuing to function reliably
    when faced with challenging conditions, such as distribution shifts, adversarial inputs, or noisy data,  
    \citep{bommasani2021opportunities,hendrycks2019using}.
\end{compactenum}

Foundation models have been applied in various fields, where they are used to solve high-level tasks in settings such as vision \citep{kirillov2023segment,liu2023grounding}, language \citep{zhou2024comprehensive}, and robotics \citep{survey-1,survey-2}. 
While robotics also deals dynamical systems, existing foundation model research has focused on \emph{multimodal} models and \emph{high-level} tasks, which can be pretrained with Internet-scale text and image data, while
leveraging general knowledge about the visual and semantic structure of the world \citep{RT-1,RT-X,VINT}. 
For instance, the RT-2 model integrates vision and language capabilities, enabling robots to interpret complex instructions and perform tasks like navigating to specific locations or manipulating objects based on visual cues. 
However, even the smallest version of RT-2, which has 5 billion parameters, can run at a frequency of around 5Hz \citep{RT-2}.
In the context of control, typical tasks such as state prediction and low-level feedback control require much faster update frequencies, and models the capture the inherent ``dynamics'' of the problem, much different from the knowledge represented in existing foundation models and available Internet-scale data.

In this work, we explore the feasibility of foundation models for dynamical systems, to be used for real-time, low-level control tasks.
We consider the state prediction problem, i.e., the ability of a model to predict future states over a horizon, which underlies many control tasks such as feedback and state estimation. 
In this context, we 
show the typical foundation model benefits \labelhighlight{P1}--\labelhighlight{P3}.
In particular, we demonstrate that a foundation model can accurately predict the future states of a dynamical system different than the one it was trained on, and that fine-tuning a foundation model with little data can yield better performance and robustness that a specialist model trained on the specific task.
Other foundation models build on the abundance of readily available datasets, e.g., from the Internet, \citep{radford2019language,mann2020language}.
However, in dynamical systems, we lack such readily available Internet-scale data for training these models. 
Thus, we ask: \emph{Can we train foundation models from purely synthetic data and still realize their benefits?}

To address these challenges, we present a novel approach that leverages the TimesFM architecture \citep{DecoderOnlyGoogle} and trains it from purely synthetic data representing dynamical systems. We propose to generate the data from dynamics functions sampled from a reproducing kernel Hilbert space (RKHS), well suited to capture a wide variety of dynamical behaviors \citep{ghojogh2021reproducingkernelhilbertspace,rosenfeld2019occupation}. 
Building on the resulting transformer-based foundation model, we investigate (and achieve) properties \labelhighlight{P1}--\labelhighlight{P3} for simulation and hardware examples. 
An overview of our proposed approach is illustrated in \Cref{fig:introduction:concept}.
To summarize, our main contributions are:
\begin{compactitem}
    \item Pretraining on purely synthetic data based on RKHS dynamics functions
    \item Validation of the foundation model claims \labelhighlight{P1}--\labelhighlight{P3} in simulation and on hardware; and
    \item Adaptation of the TimesFM architecture, including publishing our code and data.
\end{compactitem}
By showing the feasibility of foundation models for dynamical systems, this work makes an important step for leveraging the power of this recent machine learning paradigm for control.

\section{Related Work}
\label{sec:related-work}

Foundation models for dynamical systems is a new topic; except for \citet{song2024fmint,seifner2024foundational}, which were developed in parallel to ours, we are not aware of any other works in this domain. 
In contrast to \citet{song2024fmint}, which requires known dynamics and initial solutions to accelerate simulations, our approach does not assume known dynamics. 
Similarly, while \citet{seifner2024foundational} focus on imputing missing time series data governed by ODEs, our focus is forecasting multiple future states for control applications.
From a broader perspective, this work relates to three key areas: robotics, time series, and synthetic data pretraining.
Similar to foundation models in robotics, we focus on real-world applications but aim to enhance real-time control capabilities beyond high-level task execution.
This involves leveraging time series sensor data, which connects our research to time series foundation models.
However, instead of focusing on repetitive trend forecasting, we concentrate on modeling system dynamics, typically exhibiting more complex behaviors.
Additionally, we address the scarcity of large-scale real-world datasets by utilizing problem structure and expert knowledge to tailor the generation of synthetic data for pretraining.

In the domain of robotics, foundation models have predominantly concentrated on solving high-level tasks \citep{survey-1,survey-2}.
Specifically, their focus has been on leveraging text and image models to transfer general knowledge (such as object identification and understanding user text commands) to enhance robot perception.
Consequently, current foundation models for robotics adopt multimodal approaches, incorporating additional data modalities to handle Vision-Language-Action scenarios \citep{RT-X,VINT}.
Models like RT-2 exemplify this trend, processing image, text, and numerical data to facilitate exceptional perception and a more human-like understanding of the world~\citep{RT-1,RT-2}.
Similar architectures preserve tokenized data modalities, providing modularity but limiting inference speed \citep{schubert2023generalistdynamicsmodelcontrol}.
However, the high dimensionality of the encoded input renders these models computationally intensive and slow in execution, making them unsuitable for low-level, real-time control tasks.
In contrast, our work targets this limitation by developing foundation models from scratch for real-time control tasks, focusing on the ability to accurately model complex dynamical systems.

In the context of time series analysis, foundation models are predominantly applied to domains such as weather forecasting, Internet search trends, and energy usage, where the underlying patterns exhibit regular cycles and predictable behaviors.
Notable examples include TimesFM~\citep{DecoderOnlyGoogle}, Timer~\citep{liu2024timergenerativepretrainedtransformers}, and TimeGPT~\citep{timegpt}.
Current work in time series emphasizes handling stable, cyclical data, often overlooking the complexities introduced by transient dynamics that characterize more volatile or rapidly changing systems.
Additionally, most existing models focus on independent signals, disregarding the potential interactions and dependencies between multiple variables within a system \citep{PatchTST,transformers_ineffective}.
This narrow focus limits their applicability in scenarios where dynamics and interdependencies of the system are crucial.
In contrast, our work diverges by prioritizing the modeling of intrinsic system dynamics, where time-dependent changes and interactions between variables are fundamental.

In addition to advancements in robotics and time series, the pretraining of foundation models has largely depended on vast, Internet-scale datasets.
For instance, foundation models in robotics typically build upon multimodal architectures pretrained on extensive, diverse datasets~\citep{radford2019language,mann2020language}.
These models are then further trained on domain-specific datasets tailored for particular embodiments or environments.
Similarly, time series foundation models rely on large-scale datasets that aggregate data from general domains, including applications like weather forecasting and energy consumption~\citep{foundation_models,bommasani2021opportunities}.
However, the utilization of synthetic data in the pretraining process of these models remains minimal.
Existing approaches sporadically incorporate synthetic data, such as using simulations for robotics or cyclic functions for time series models \citep{DecoderOnlyGoogle,RT-X}.
Despite these efforts, the predominant reliance is still on preexisting, real-world Internet-scale datasets for initial pretraining, not available for low-level control and dynamical systems.
In contrast, our work focuses on the use of purely synthetic data for training foundation models, leveraging data generation from RKHS dynamics functions, which has not been explored before.

\section{Problem Formulation}

We aim to model the dynamics of discrete-time dynamical systems, defined by the transition function \(f: \mathcal{X} \times \mathcal{U} \rightarrow \mathcal{X}\), where the state \(x_{k+1}\) evolves as
\begin{equation}
    x_{k+1} = f(x_k, u_k) + \varepsilon_k,
\end{equation}
with process noise \(\varepsilon_k\).
The state space \(\mathcal{X} \subseteq \mathbb{R}^{d_x}\) and action space \(\mathcal{U} \subseteq \mathbb{R}^{d_u}\) represent the system's state and control input dimensions, respectively.
We operate on finite-length trajectories consisting of state-action pairs, denoted as \(\mathbf{x} = \{(x_1, u_1), \dots, (x_T, u_T)\}\).
Subtrajectories are expressed using slicing notation, e.g., \(\mathbf{x}_{k:k+H}\), representing the sequence from time step \(k\) to \(k+H\).
This notation extends to sequences of only states or only actions, such as \(x_{k:k+H}\) and \(u_{k:k+H}\).

The goal is to train a foundation model \(\mathcal{F}_\theta\) that generalizes across a family of dynamical systems that are randomly drawn from the space \(\mathcal{H}\).
For any \(f \in \mathcal{H}\), the model should precisely predict future states based on past states and actions.
Given a context window of length \(c \in \mathbb{N}\) and a prediction horizon of length \(m \in \mathbb{N}\), the model predicts the future states \(\hat{x}_{k+1:k+m}\) using the context \(\mathbf{x}_{k-c:k-1}\) and future actions \(u_{k:k+m-1}\):
\begin{equation}
    \hat{x}_{k+1:k+m} = \mathcal{F}_\theta(\mathbf{x}_{k-c:k-1}, u_{k:k+m-1}).
    \label{eq:state-prediction}
\end{equation}
The objective is to minimize the prediction error for all \(f \in \mathcal{H}\), ensuring robust and accurate performance across diverse dynamical systems.
This state prediction problem essentially tests the capabilities of the model $\mathcal{F}_\theta$ to capture the relevant dynamics of a system, which is needed in basically all of model-based control. Explicitly, state prediction like (\ref{eq:state-prediction}) is relevant in model-predictive control or state estimation, for example.

\paragraph{Evaluation Objectives}
We consider the evaluation objectives \labelhighlight{P1}--\labelhighlight{P3}.
We assess the model's ability to generalize by evaluating its mean squared error across a range of dynamical systems, including both seen and unseen systems from the family \(\mathcal{H}\).
The model's data efficiency is quantified by its performance, across varying amounts of fine-tuning data.
A data-efficient model achieves strong performance even when fine-tuned with minimal data or in zero-shot predictions (i.e., when the pretrained foundation model is used directly without any find-tuning). 
Robustness is measured by the consistency of the model’s performance across different training runs and fine-tuning configurations.
It is evaluated by analyzing the range of MSE values across multiple runs with different initializations and data subsets.

\section{Synthetic Data Generation}
\label{sec:synthetic-data-generation}

The goal of our synthetic data generation is to obtain representative data that resemble general dynamical systems. In particular, we need a function space that is large enough, has easily adjustable properties such as smoothness, and can efficiently be sampled from.
We propose to sample dynamics functions from an RKHS to create trajectory data. 
Kernel methods and RKHSs are frequently used in the context of dynamical systems \citep{pillonetto2014kernel, care2023kernel}. Additionally, they have close connections to Gaussian process regression \citep{williams2006gaussian, kanagawa2018gaussian}, which is often used to learn and represent the dynamics function \citep{buisson2020actively, treven2023optimistic, svensson2017flexible, geist2020learning}. 
Among other advantageous properties such as data efficiency and embedding of prior knowledge, uniform error bounds are tractable \citep{srinivas2012information, fiedler2021practical}.

Whenever we represent dynamics functions with kernel methods, we obtain the approximation \begin{equation}
    f(x) = \sum_{i=1}^{n} \alpha_{i} k(x, x_{i}),
    \label{eq:RKHSfunctions}
\end{equation}
where \(x_{i}\) are supporting (data) points and \(\alpha_{i}\) coefficients obtained from the specific training method.
 There is a one-to-one correspondence between each RKHS $\mathcal{H}$ and the corresponding kernel function $k(.,.)$. 
Further, the norm of \(f\) in the RKHS \(\mathcal{H}\) is given by \(\|f\|_{\mathcal{H}}^2 = \sum_{i=1}^{n} \sum_{j=1}^{n} \alpha_i \alpha_j k(x_i, x_j)\).

We use \cref{eq:RKHSfunctions} to sample functions from the RKHS. 
As a first design choice, we focus on dynamical systems without any control inputs. The control is reintroduced later again in Sec.\,\ref{Sec:experiments}. 
Further, we break down the problem of sampling the vectorfield \(f \) into sampling \(n\) independent functions \(f_i: \mathbb{R}^{d_x} \rightarrow \mathbb{R}\), where \(f(x) = (f_1(x), \dots, f_n(x))^\top\). We choose the standard RBF kernel \(k(x, x') = \sigma^2\exp(-\frac{\|x - x'\|^2}{2l^2})\) with hyperparameters $\sigma^2$ and $l$.
We sample the supporting points \(x_{i}\) uniformly from \([x_{\min}, x_{\max}]^{d_x}\) and coefficients \(\alpha_{i} \sim \mathcal{N}(0, \sigma_\alpha^2)\) randomly, where \(x_{\min}, x_{\max}, \sigma_\alpha^2\) are hyperparameters.
To generate functions that represent a broad variety of behaviors, we scale the RKHS norm of each function to a target value selected from a uniform distribution between \(\text{norm}_{\min}\) and \(\text{norm}_{\max}\), enabling to sample functions in a specific norm range. We scale the coefficients \(\alpha_i\) to ensure the function achieves the target norm \(\alpha_i' = \alpha_i \cdot \frac{\text{norm}_{\text{target}}}{\|f\|_{\mathcal{H}}}\). 
Our approach of sampling RKHS functions is consistent with \citep{fiedler2021practical} and \citep{steinwart2008support}.

\begin{figure}[tb]
    \centering
    \includegraphics[width=0.6\textwidth]{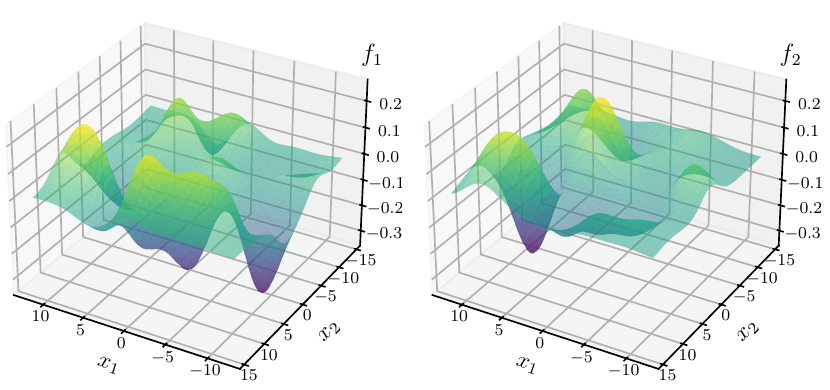}
    \includegraphics[width=0.3\textwidth]{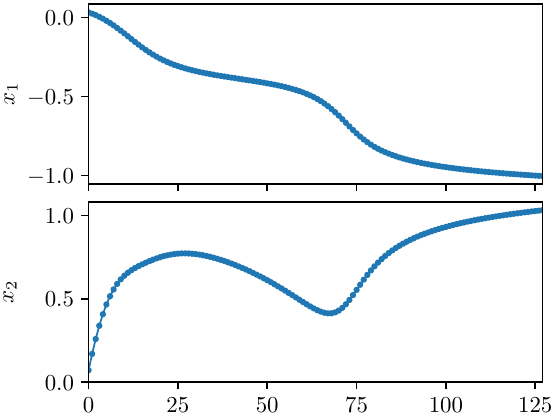}
    \caption[Example Dynamics and Trajectory sampled from an RKHS.]{Representative functions sampled from an RKHS (left) in a two-dimensional state (\(x_1, x_2\)) space and an example trajectory generated from the sampled dynamics functions (right).}
    \label{fig:data:example_dynamics_trajectory}
\end{figure}

We consider the sampled functions $f$ to correspond to a continuous time dynamical system. This way, we have more control over smoothness and discretization properties.
To obtain trajectories, we use Euler’s method to discretize the system dynamics, where the next state is given by \(x_{k+1} = x_k + \Delta t \cdot f(x_k)\) and we choose the discretization step \(\Delta t\) as a hyperparameter. In principle, any other method such as Runge-Kutta approaches could be used here as well.

Depending on the sampled dynamics and initial point $x_0$, data might look extremely similar. 
Thus, we propose a total variation selection criterion to remove overrepresented behavior and select trajectories, where the model can learn a meaningful general representation.
The total variation of a trajectory \(\mathbf{x} = \{x_0, x_1, \dots, x_T\}\) is defined~\citep{royden2010real} as $  \mathrm{TV}(\mathbf{x}) = \sum_{k=1}^{T} \|x_k - x_{k-1}\|$. We select only those trajectories whose total variation \(\mathrm{TV}(\mathbf{x})\) falls within the predefined bounds \(\mathrm{TV}_{\min} \leq \mathrm{TV}(\mathbf{x}) \leq \mathrm{TV}_{\max}\). 
Furthermore, we split the trajectories into context part \(\mathbf{x}_{\text{context}} = \mathbf{x}_{k-c:k-1}\) and predicted part \(\mathbf{x}_{\text{predicted}} = \mathbf{x}_{k+1:k+m}\) to select trajectories based on the total variation difference between the two halves.
We discard trajectories, where \(|\mathrm{TV}(\mathbf{x}_{\text{context}}) - \mathrm{TV}(\mathbf{x}_{\text{predicted}})| > \delta\).
The underlying assumption is that the input and output parts of the trajectory should exhibit similar dynamics and thus have similar total variation.
In practice, we choose \(\delta\) slightly smaller than \(\mathrm{TV}_{\max} - \mathrm{TV}_{\min}\) to discard trajectories where the output part is nearly constant.
Finally, we bin the trajectories based on their total variation and downsample from overrepresented bins to ensure a balanced distribution of trajectories across the total variation spectrum.
A representative function sample with its corresponding trajectory are shown in \Cref{fig:data:example_dynamics_trajectory}.

We summarize our approach: (i) sample $N$ dynamics functions from the RKHS, (ii) scale each dynamics function's coefficients to a selected norm, (iii) sample a single trajectory from each dynamics functions using Euler's method, yielding \(N\) trajectories, (iv) select trajectories based on a total variation threshold and the total variation threshold of the context and prediction part, and (v) bin trajectories based on total variation and downsample overrepresented bins.

\section{Model}
\label{sec:model}

We utilize a decoder-only transformer model, as introduced by \citet{DecoderOnlyGoogle}, which achieves state-of-the-art performance for time series foundation models. Specifically designed for time series forecasting, this architecture can effectively capture complex temporal dependencies and nonlinear dynamics, making it ideal for modeling dynamical systems.

\paragraph{Model Architecture} The model processes sequences of state-action pairs, \((x_k, u_k) \in \mathcal{X} \times \mathcal{U}\).
Given an input trajectory \(\{(x_1, u_1), \dots, (x_T, u_T)\}\), it predicts a future trajectory \(\{\hat{x}_2, \dots, \hat{x}_{T+1}\}\). 
Only predictions (\(\{\hat{x}_{c+1}, \dots, \hat{x}_{c+m}\}\)) beyond the observed context are used for evaluation. 
Formally, given
\begin{equation}
    (\mathbf{x}_{\text{input}}, \mathbf{u}_{\text{input}}) = \{(x_1, u_1), \dots, (x_c, u_c), (0, u_{c+1}), \dots, (0, u_{c+m})\},
\end{equation}
where states beyond \(c\) are set to zero as placeholders, the model generates \(\{\hat{x}_2, \dots, \hat{x}_{c+m+1}\}\), using only \(\{\hat{x}_{c+1}, \dots, \hat{x}_{c+m}\}\) as valid predictions for the true future states.

\begin{figure}[tb]
    \centering
    \includegraphics[width=0.8\textwidth]{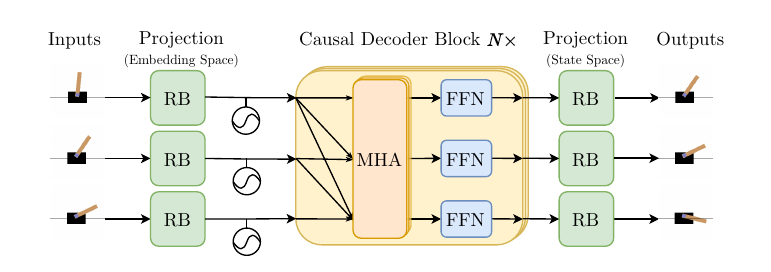}
    \caption[Model Architecture]{Architecture of the decoder-only transformer model.
    The model projects inputs into an embedding space using a residual block (RB), followed by $N$ causal decoder blocks with multi-head self-attention (MHA) and feed-forward networks (FFN).
    A final residual block projects from the embedding space to the output space.}
    \label{fig:model:overview}
\end{figure}

The overall model architecture is illustrated in \Cref{fig:model:overview}. 
Input state-action pairs \((x_i, u_i)\) are projected into a high-dimensional embedding space via a Residual Block~\citep{residualblock}, expressed as \(x_{i,\text{embed}} = \text{ResidualBlock}(x_i, u_i)\), capturing intricate system dynamics.
The embedded sequence is then processed by a causal decoder with \(N\) decoder blocks using self-attention~\citep{attentionisallyouneed}.
This causal design ensures predictions depend only on past and present inputs, enabling the capture of long-range dependencies to predict future states:
\begin{equation}
    x_{0:c+m, \text{decoder}} =  \text{decoder}(x_{0:c+m, \text{embed}}).
\end{equation}

\paragraph{Model Training} The model undergoes a two-phase training process: \textbf{pretraining} on synthetic data, as described in \Cref{sec:synthetic-data-generation}, to generalize across various dynamical systems, followed by optional \textbf{fine-tuning} on task-specific datasets using a reduced learning rate and fewer epochs for system-specific adaptation.
We optimize the model using the AdamW optimizer~\citep{loshchilov2019decoupledweightdecayregularization}, which combines adaptive learning rates with weight decay for stable and efficient training.
Following \citet{DecoderOnlyGoogle}, we increase the model's flexibility and robustness by employing random masking (masking 10\% of input states) and output patching with a patch size of two while omitting input patching. 
This enables the model to predict two future states simultaneously during training, and focus on the actual dynamics of the series.

\paragraph{Data Augmentation} We apply data augmentation techniques to improve model training and generalization, distinguishing between synthetic data during pretraining and dynamical systems data during fine-tuning.
\textbf{During pretraining}, we apply scaling and shifting~\citep{wen2020time} to trajectories:
\(
\mathbf{x}' = \alpha \mathbf{x} + \beta,
\)
where \(\alpha \in [\alpha_{\min}, \alpha_{\max}]\) and \(\beta \in [\beta_{\min}, \beta_{\max}]\) are randomly sampled from uniform distributions.
\textbf{During fine-tuning}, we add Gaussian noise (iid) to each state measurement of the dynamical systems data:
\(
x_k' = x_k + \varepsilon_k,
\)
where \(\varepsilon_k \sim \mathcal{N}(0, \sigma^2)\).

\section{Experiments}
\label{Sec:experiments}
In this section, we will validate that pretraining our transformer model on RKHS data yields the typical foundation model properties \labelhighlight{P1}--\labelhighlight{P3}.\footnote{The code will be released upon acceptance.} In particular, we will show:  (i) our model generalizes, i.e., it is able to accurately predict future states of completely unseen systems, indicating that the transformer architecture is a good choice overall; (ii) learning is more robust, i.e., our pretrained transformer model performs best and additionally shows significantly lower variance than all other baselines; and
(iii) we are more data efficient, i.e., the loss of the pretrained model is substantially lower than of all other models when new data is used for fine-tuning instead of learning from scratch.

We evaluate the proposed pipeline on both simulation and hardware systems.
Our transformer model, comprising 20 layers and approximately 3.4 million parameters, is pretrained on synthetic RKHS data and optionally fine-tuned on dynamical systems data.
This yields two models: the \textbf{pretrained model (Pre)} and the \textbf{fine-tuned model (Ft)}, which are evaluated on test sets and compared against baselines.
Each experiment is repeated 20 times for statistical significance.

\begin{wrapfigure}[20]{r}{0.3\textwidth}
    \vspace{-1em} 
    \centering
    \begin{minipage}{\linewidth}
        \centering
        \includegraphics[width=0.9\linewidth]{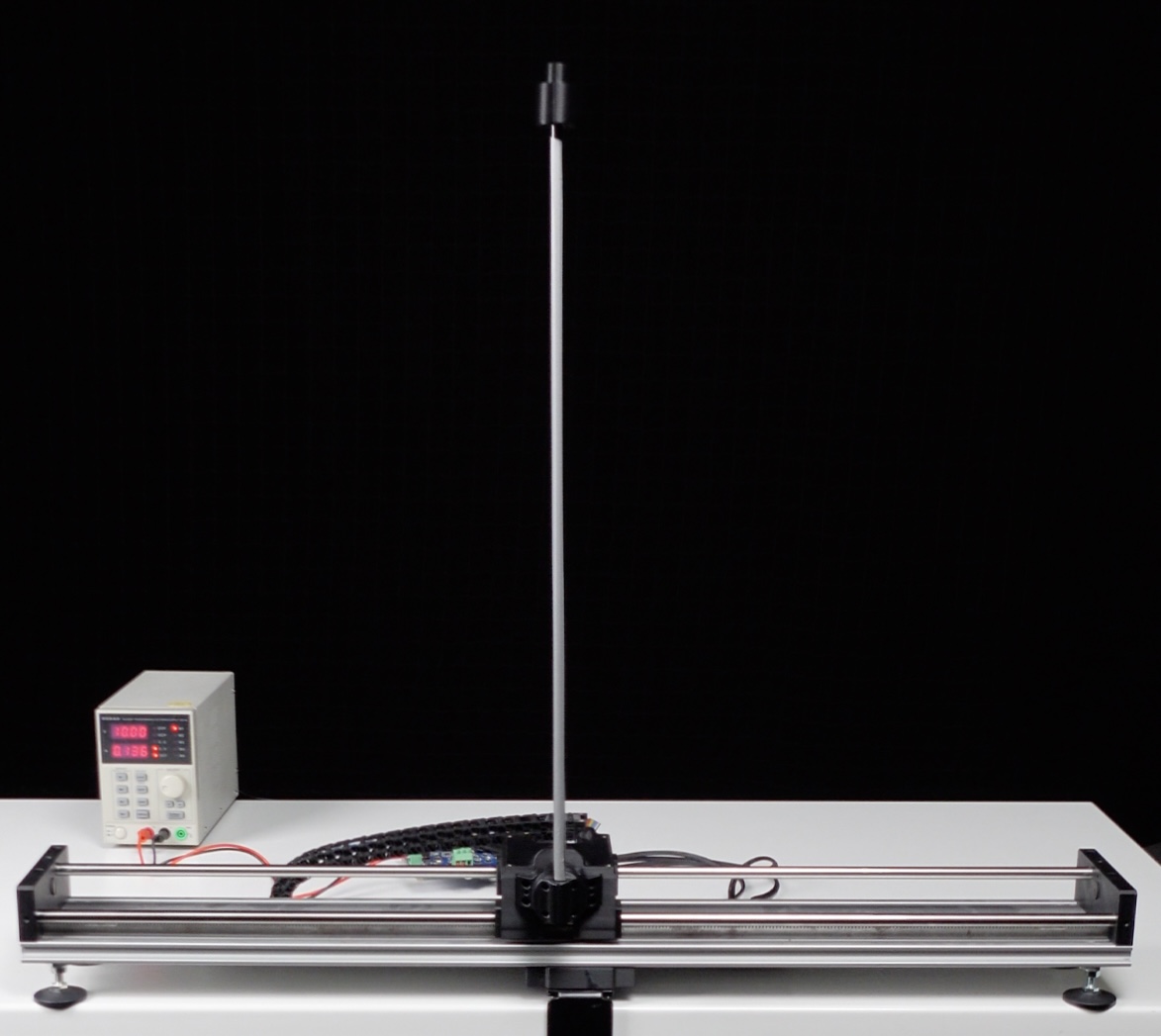}
    \end{minipage}
    \vspace{0.5em}
    \begin{minipage}{\linewidth}
        \centering
        \includegraphics[width=0.82\linewidth]{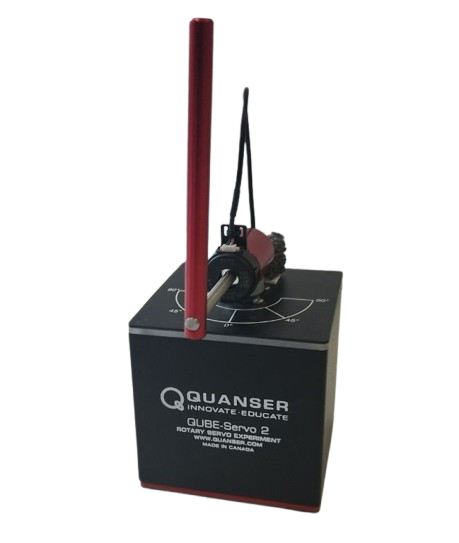}
    \end{minipage}
    \vspace{-0.6cm}
    \caption{Custom cart-pole (image taken from \citep{hose2024parameteradaptiveapproximatempctuning}) and Quanser Qube Servo 2 Furuta systems.}
    \label{fig:experiments:hardware_systems}
\end{wrapfigure}

\paragraph{Datasets}
We use both simulated and real-world datasets for training and evaluation.
The \textit{simulated datasets} are:
(i) a cart-pole simulation with fixed parameters and constant action of zero, where the pole is randomly initialized near the upright position and swings freely, and
(ii) a cart-pole simulation with randomized parameters and pink noise action, with each trajectory reflecting a unique system configuration.
The \textit{hardware datasets} are:
(iii) a custom cart-pole system with action, whose parameters are described in \citep{hose2024parameteradaptiveapproximatempctuning}, and
(iv) a Furuta pendulum using the Quanser Qube Servo 2~\citep{QuanserQube2024}.
Both hardware systems are shown in Figure~\ref{fig:experiments:hardware_systems}.

\subsection{Experimental Setup}
\paragraph{Evaluation Metrics}
We evaluate our models based on the defined objectives \labelhighlight{P1}--\labelhighlight{P3}.
The evaluation metric for generalization is the mean squared error (MSE) over the trajectory.
For the unseen part of the trajectory (i.e., the prediction horizon) of length \(m\), we calculate the MSE between the predicted states \(\hat{\mathbf{x}}_t\) and true states \(\mathbf{x}_t\) as \(\text{MSE} = \frac{1}{m} \sum_{t=1}^{m} \left\| \hat{\mathbf{x}}_t - \mathbf{x}_t \right\|^2\).
Furthermore, we assess the robustness of the model by evaluating the range of MSE values across the worst and best runs.
A smaller range indicates a more stable and robust model.
To investigate data efficiency, we conduct experiments using datasets of varying sizes: 2\%, 10\%, and 100\% of the total 20,000 trajectories for each training dataset and compare the MSE across these data levels.
In this setup, the pretrained model is treated as having seen 0\% of the data, as it has not seen any data from the cart-pole system.
We compare the performance of our proposed model against several baselines:
\begin{compactitem}
    \item \textbf{Linear Regression (LR)}: A classical method with limitations in capturing nonlinearity.
    \item \textbf{Feedforward Neural Network (FNN)}: A small neural network with three layers consisting of 128, 64, and 32 neurons, respectively, using ReLU activation functions.
    \item \textbf{Smaller Transformer (ST)}: A reduced version of the proposed transformer model, using the same decoder-only architecture but with 8 layers and approximately 200k parameters.
    This baseline is not pretrained and thus shows the impact of pretraining and the foundation model paradigm.
    This model learns only task-specific representations, allowing it to be smaller.
\end{compactitem}
Since LR and FNN lack the causality mechanism inherent in transformer architectures, we apply an iterative prediction approach.
In this setup, the predicted state is fed back into the model to predict the subsequent state, resulting in the prediction formulation
$\mathbf{x}_{k:k+31} \leadsto \hat{x}_{k+32}^{\text{LR}}$, where the model takes state-action pairs from steps \(k\) to \(k+31\) as input to predict the state at step \(k+32\).
This ensures a fair comparison, since all models have access to the same data for each prediction step.

\subsection{Simulation Cart-Pole Experiments}

\begin{figure}[tb]
    \centering
    \includegraphics[width=0.66\textwidth]{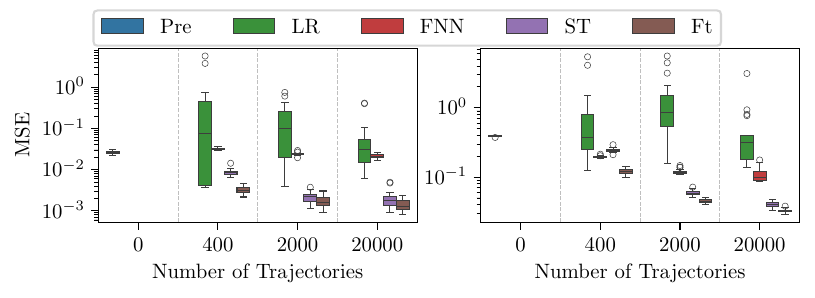}
    \includegraphics[width=0.32\textwidth]{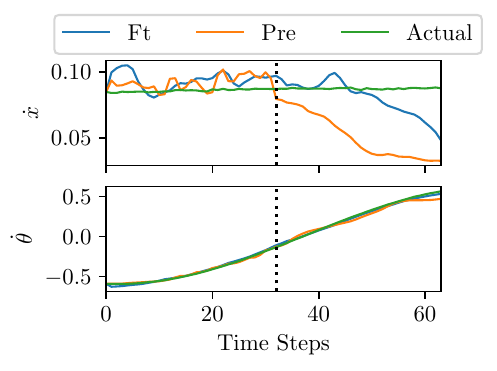}
    \caption{Simulation results for cart-pole with fixed parameters and constant action (left) and randomized parameters with pink noise action (middle).
    We compare the MSE of our models (Pre, Ft) against baselines (LR, FNN, ST) across data subsets, showing generalization, data efficiency, and robustness (range of MSE values).
    The right shows 
    an example trajectory, comparing the pretrained and fine-tuned models with the actual elocities \(\dot{x}\) and \(\dot{\theta}\) (note the different scaling of y-axes).}
    \label{fig:results:cartpole_constant_fixed_and_random_pink}
\end{figure}

The simulation results are shown in \Cref{fig:results:cartpole_constant_fixed_and_random_pink}.
Across both experiments, the transformer models show good generalization capabilities, indicated by a low MSE. 
Furthermore, the fine-tuned model outperforms the baselines, especially the smaller transformer, showing the generalization capabilities and the advantages of pretraining on synthetic data.
This shows that the pretraining process significantly enhances the fine-tuning phase, enabling faster convergence and better performance compared to the smaller transformer, especially in low-data scenarios.

Furthermore, the pretrained transformer model achieves reasonable performance despite not having seen any data from the cart-pole system.
In the fixed parameter experiment without actions, it outperforms the linear regression baseline and performs comparably to the feedforward neural network in low-data scenarios.
This highlights the model's effectiveness in zero-shot prediction scenarios, where it can generalize to unseen systems without fine-tuning.

In the randomized parameters experiment with pink noise, the pretrained transformer model shows reasonable performance, though accuracy is reduced compared to the fixed parameters case.
This decline is likely due to random actions introducing unseen dynamics, which the model did not encounter during pretraining.
Nevertheless, the pretrained model provides a robust initialization, contributing to the fine-tuned model's superior performance and robustness across conditions.

\Cref{fig:results:cartpole_constant_fixed_and_random_pink} (right) shows a specific cart-pole trajectory with fixed parameters and constant action, comparing the pretrained and fine-tuned models against the actual trajectory (for the velocity \(\dot{x}\) and angular velocity \(\dot{\theta}\)).
This trajectory shows the strength of the pretrained model and the improvement after fine-tuning, as the fine-tuned model more closely follows the actual trajectory.

\subsection{Hardware Cart-Pole and Furuta Experiments}

\begin{figure}[tb]
    \centering
    \includegraphics[width=0.66\textwidth]{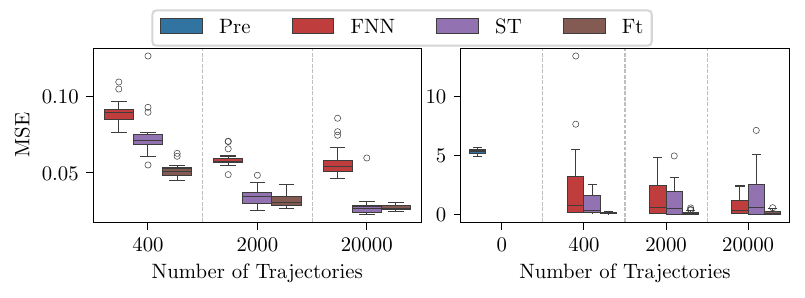}
    \includegraphics[width=0.32\textwidth]{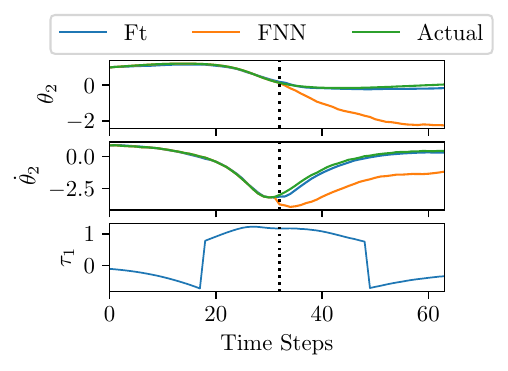}
    \caption{Hardware results for the cart-pole (left) and Furuta pendulum (middle).
    We compare the MSE of our models (Pre, Ft) against baselines (FNN, ST) across data subsets, showing generalization, data efficiency, and robustness (range of MSE values).
    The right plot shows a hardware Furuta trajectory, comparing the fine-tuned model, FNN baseline, and the actual arm angle \(\theta_2\) and velocity \(\dot{\theta}_2\), with control action \(\tau_1\).}
    \label{fig:results:hardware_cartpole_and_furuta}
\end{figure}

Results from applying the foundation model on actual experimental data 
are shown in \Cref{fig:results:hardware_cartpole_and_furuta}.
We exclude linear regression due to its bad performance, showing Ft and Pre for our model. Zero-shot performance (Pre) is not competitive here, which is not surprising as pretraining considered zero action ($u \equiv 0$). Action is highly relevant in the experiments, and Ft can adapt to it.
While the hardware experiments confirm the simulation results, we highlight additional insights next.

For the hardware cart-pole and the full dataset (20,000 trajectories), the best run of the smaller transformer slightly outperforms the fine-tuned transformer, indicating that a large amount of data can partially offset the benefits of pretraining and increased model capacity.
However, the median performance of the fine-tuned transformer remains superior, underscoring its reliability and data-efficiency.
Adding to these results, the fine-tuned transformer exhibits robust performance, as evidenced by its narrow range of MSE values across runs.

The results for the hardware Furuta pendulum system follow similar trends.
While all baseline models exhibit a high range of MSE values across runs, the fine-tuned transformer consistently shows a narrow range, reflecting its stable and reliable performance.
Adding to the findings, this robustness is maintained across all data subsets, suggesting that increasing data alone does not resolve the variability issues observed in the baselines.

\Cref{fig:results:hardware_cartpole_and_furuta} (right) shows a specific trajectory of the hardware Furuta system, comparing the fine-tuned model against the FNN baseline and the actual trajectory.
We selected the arm angle \(\theta_2\) and arm angular velocity \(\dot{\theta}_2\) as representative examples.
This shows the strength of the fine-tuned model, which closely follows the actual trajectory and outperforms the FNN baseline.

\section{Conclusion}

In this work, we investigated and validated the viability of foundation models for low-level state predictions of dynamical systems.
Our long-term vision is the design of one generalist model that is capable of real-time control for a large class of dynamical systems.
As a first step, we have shown in this paper that the idea and typical advantages of foundation models apply to dynamical systems and specifically the state-prediction problem.
Our training pipeline uses purely synthetic data, avoiding the need for a giant and currently unavailable real-world dataset.
The transformer-based model is pretrained and generalizes well to unseen simulation and real-world systems.
Through fine-tuning to a specific system, we can increase performance with little data even further.
In particular, we demonstrate that our model outperforms common baselines in the defined foundation model properties \labelhighlight{P1}--\labelhighlight{P3}.

\acks{
We would like to thank Henrik Hose, Alexander Gräfe, Paul Brunzema, and Sebastian Giedyk for their support with the custom cart-pole system and the Furuta pendulum, as well as for the insightful discussions.
In addition, model training was partially performed with computing resources granted by RWTH Aachen University, under project \texttt{thes1658}.

}

\bibliography{reference}

\end{document}